\documentclass[sigconf]{acmart}
\AtBeginDocument{%
  }

\usepackage{colortbl}
\usepackage{algorithm}
\usepackage{algorithmic}
\usepackage{booktabs}
\usepackage{listings}
\usepackage{float}
\usepackage{colortbl} 
\usepackage{multirow}
\usepackage{tcolorbox}
\usepackage{multicol}
\usepackage{balance}
\usepackage{xcolor} 
\usepackage{hyperref}

\acmYear{2025}\copyrightyear{2025}
\setcopyright{acmlicensed}
\acmConference[FSE Companion '25]{33rd ACM International Conference on the Foundations of Software Engineering}{June 23--28, 2025}{Trondheim, Norway}
\acmBooktitle{33rd ACM International Conference on the Foundations of Software Engineering (FSE Companion '25), June 23--28, 2025, Trondheim, Norway}
\acmDOI{10.1145/3696630.3728522}
\acmISBN{979-8-4007-1276-0/25/06}

\begin{document}

\title{Can Hessian-Based Insights Support Fault Diagnosis in Attention-based Models?}

\author{Sigma Jahan}
\affiliation{
    \institution{Dalhousie University}
    \country{Canada}
}
\email{sigma.jahan@dal.ca}

\author{Mohammad Masudur Rahman}
\affiliation{
    \institution{Dalhousie University}
    \country{Canada}
}
\email{masud.rahman@dal.ca}

\begin{abstract}
As attention-based deep learning models scale in size and complexity, diagnosing their faults becomes increasingly challenging. In this work, we conduct an empirical study to evaluate the potential of Hessian-based analysis for diagnosing faults in attention-based models. Specifically, we use Hessian-derived insights to identify fragile regions (via curvature analysis) and parameter interdependencies (via parameter interaction analysis) within attention mechanisms. Through experiments on three diverse models (HAN, 3D-CNN, DistilBERT), we show that Hessian-based metrics can localize instability and pinpoint fault sources more effectively than gradients alone. Our empirical findings suggest that these metrics could significantly improve fault diagnosis in complex neural architectures, potentially improving software debugging practices.
\end{abstract}



\maketitle

\section{Introduction}
Deep learning models with attention mechanisms have become integral in a variety of tasks, including natural language processing~\cite{devlin2019bert, vaswani2017attention}, recommendation systems~\cite{zhang2019deep}, and code understanding~\cite{yang2022survey}. Their ability to selectively focus on different parts of an input sequence (i.e., attention mechanism~\cite{vaswani2017attention}) has significantly improved their accuracy and versatility compared to previous networks (e.g., RNNs). In particular, transformer models adopting self-attention mechanisms, including the GPT series~\cite{liu2024gpt} and BERT~\cite{devlin2019bert}, have become foundational in many application domains. As DL models advance in scale and complexity, the loss function remains a key factor in guiding optimization and determining whether training converges smoothly or encounters instability~\cite{goodfellow2016deep}. Earlier architectures (e.g., CNNs, RNNs) often exhibit smoother loss surfaces that can guide their model optimization through a more predicted path~\cite{goodfellow2015qualitatively, li2018visualizing}. On the other hand, attention-based models handle more complex tasks and diverse data types by adopting larger and more intricate architectures. Although they may perform well for specific complex tasks, their increased complexity often results in a loss surface that is more complex or irregular, which can challenge their optimization~\cite{hao2021visually, He2020Piecewise, wang2021deep}. They could also suffer from unstable training, slow convergence, or failures during fine-tuning~\cite{dehghani2023scaling, wortsman2024small, zhai2023stabilizing, li2020scaling}. 

Faults in attention-based models often stem from their unique architectures, where multiple attention heads (with query, key, and value weights) interact with each other in diverse, non-linear ways to decide which parts of an input matter most~\cite{vaswani2017attention}. Even a small update to the query, key, or value parameters might drastically change attention patterns~\cite{dong2021attention}, leading to training instabilities~\cite{zhai2023stabilizing}. Moreover, changing a single attention head can cascade through others, potentially altering attention behaviour~\cite{attentionExplanation, attentionInterpretable}. For example, in a translation model, if one head misplaces its focus on the word \textit{play}, other heads might incorrectly emphasize related words like \textit{music} or \textit{stage}. As a result, the overall translation might shift from sports to music. Such cascading effects might also lead to \textit{attention entropy collapse}~\cite{dong2021attention, zhai2023stabilizing}.

Given the huge parameter space, these subtle and interconnected behaviors make faults in attention-based models harder to diagnose, underscoring the need for more robust diagnostic methods. Many techniques have leveraged gradient-related information to diagnose performance-related faults (e.g., vanishing gradient) in deep neural networks (e.g., CNNs, RNNs)~\cite{wardat2021deeplocalize, zhang2021autotrainer, deepfd, umlaut}. Gradients measure how a single parameter of a model affects the overall loss (i.e., global loss)~\cite{goodfellow2014explaining}. However, However, they might
fall short for more complex networks such as attention-based networks, since they cannot effectively capture multi-parameter interactions~\cite{yao2018hessian}.

Recent studies~\cite{serrano2019attention, singh2021analytic, ghorbani2019investigation} suggest that second-order derivatives (i.e., Hessians) offer a promising direction for addressing these limitations. By capturing how gradients change in the local neighborhood of a parameter, Hessians provide deeper insights into the loss surface (i.e., local loss), potentially identifying \emph{unstable regions}. This is particularly relevant in attention-based models, where small parameter updates can abruptly alter attention patterns. Furthermore, Hessians capture how changes in one parameter influence others (i.e., multi-parameter interaction), which could reveal the source of fault propagation across attention heads. These properties highlight the potential of Hessians for revealing hidden faults in attention-based models. Although Hessian-based analysis has been explored in CNNs and RNNs~\cite{jiang2019fantastic, martens2010deep, yao2018hessian}, its potential remains largely unexplored in the context of attention-based models. 
\begin{figure*}[h!]
    \centering
    \vspace{-1.0em}
    \includegraphics[width=0.9\linewidth]{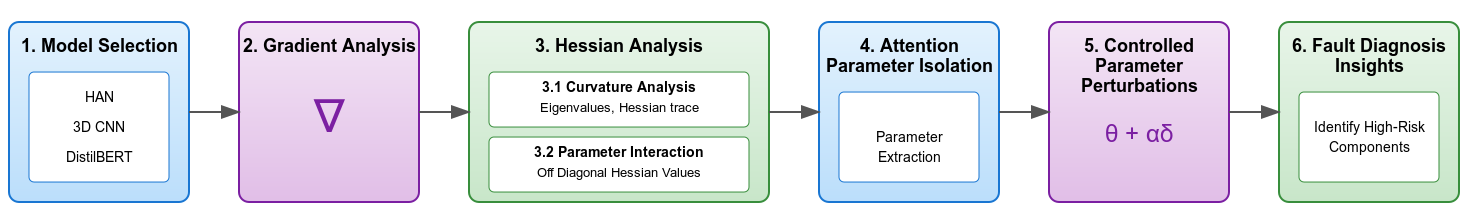}
    \vspace{-1.0em}
    \caption{Schematic Diagram of the Study}
    \label{fig:diagram_study}
    \vspace{-1.5em}
\end{figure*}

In this paper, we address this gap through an \emph{empirical study}, applying Hessian-based analysis to attention-based models (e.g., DistilBERT, HAN) to explore its effectiveness in diagnosing faults (e.g., training instability, performance issues). Thus, we ask two research questions as follows.\\
\indent
\textbf{RQ$_1$: Can we identify unstable regions in attention-based models that may lead to training issues?}
\begin{tcolorbox}[colback=gray!10, colframe=white, arc=2mm, boxrule=0mm,
    boxsep=4pt, left=2pt, right=2pt, top=2pt, bottom=2pt]
We found that analyzing the local loss surface (i.e., curvature analysis) using Hessians revealed regions with strong signs of instability (Tr(\textbf{H}) $-5.72$), missed by gradients alone, and identified specific attention parameters responsible.
\end{tcolorbox}

\textbf{RQ$_2$: Can we understand how faults propagate across components in attention-based models?}
\begin{tcolorbox}[colback=gray!10, colframe=white, arc=2mm, boxrule=0mm,
    boxsep=4pt, left=2pt, right=2pt, top=2pt, bottom=2pt]
We observed that analyzing the parameter interactions using Hessians revealed strong correlations (e.g., $-0.68$) among attention components, clearly demonstrating how faults might propagate and cascade through the attention mechanisms.
\end{tcolorbox}

\vspace{-1.0em}
\section{Methodology}
\label{sec:methodology}

\subsection{Selection of Attention-based Models}
\label{subsec:data_collection}

Fig.~\ref{fig:diagram_study} shows the schematic diagram of our conducted study. We discuss the major steps of our study design as follows.

To conduct our study, we carefully selected three attention-based models -- HAN, 3D-CNN, and DistilBERT. Each model adopts a distinct architecture and targets a particular task. First, the Hierarchical Attention Network (HAN) for document classification\cite{HAN} combines recurrent neural networks with multilevel self-attention. Second, the 3D CNN architecture employs cross-attention for video classification tasks \cite{pytorch_text_classification}. Lastly, a fine-tuned DistilBERT model for sentiment analysis targets the built-in multi-head self-attention mechanism~\cite{huggingface_transformers}. These models cover a range of modalities (e.g., text, video, and sentiment) and attention designs (e.g., multilevel, cross, and multi-head self-attention). We obtained their original implementations from their respective open-source repositories~\cite{pytorch_text_classification, hmdb51_recognition, huggingface_transformers} and applied necessary instrumentation to them for gradient and Hessian-based analyses.

\vspace{-0.5em}
\subsection{Gradient Computation}
Gradient-based methods rely on first-order derivatives of the loss function ($L$, where $L = \mathcal{L}(\theta)$) to identify which parameters ($\theta = \{\theta_1, \theta_2, \ldots, \theta_n\}$) have the strongest immediate effect on the loss~\cite{GradientVsHessian_theory}. We use PyTorch’s automatic differentiation \cite{pytorch_autograd_tutorial} to measure gradient norms of the model parameters. 
\begin{figure*}[h]
    \centering
        \vspace{-1.0em}
    \includegraphics[width=0.9\linewidth]{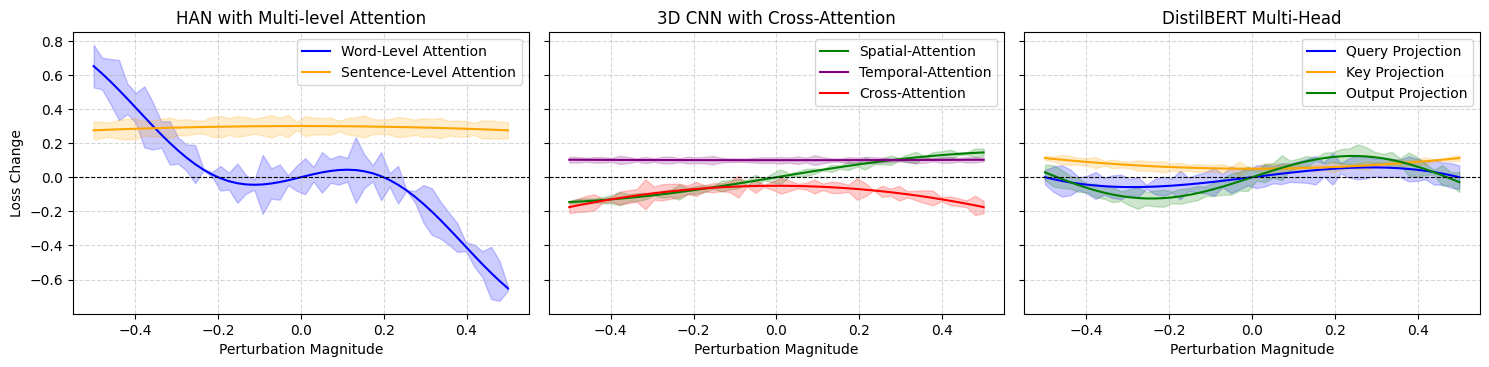}
        \vspace{-1.0em}
    \caption{Demonstration of Loss Sensitivity to Parameter Perturbations}
            \label{fig:loss_pertubation}
        \vspace{-1.0em}
\end{figure*}

\begin{table*}[h]
    \centering
    \caption{Curvature Analysis of Attention Mechanisms Across Models}
    \vspace{-1.0em}
    \small
    \label{tab:curvature_analysis}
    \resizebox{0.9\textwidth}{!}{ 
    \renewcommand{\arraystretch}{0.90} 
    \setlength{\tabcolsep}{3.5pt} 
    \begin{small} 
    \begin{tabular}{|l|l|l|l|p{7.5cm}|}
        \hline
        \textbf{Model} & \textbf{Layer} & \textbf{H.Trace} & \textbf{Top Eigenvalues} & \textbf{Interpretation} \\
        \hline
        \multirow{2}{*}{HAN} 
        & Word Attention & \cellcolor{red!20}\(-5.72\) & \((-0.4186, -0.0007, 0.4493)\) & Concave; highly unstable \& sensitive to parameter changes~\cite{martens2015optimizing}. \\
        & Sentence Attention & \cellcolor{green!20}\(2.86\) & \((-0.0007, 0.0061, 0.0387)\) & Convex; stable \& less prone to training failures~\cite{cohen2021gradient}. \\
        \hline
        \multirow{3}{*}{3D CNN} 
        & Spatial Attention & \cellcolor{green!20}\(1.92\) & \((-0.0004, 0.0103, 0.0228)\) & Convex; stable \& supports smooth optimization~\cite{jastrzebski2020break}. \\
        & Temporal Attention & \cellcolor{green!20}\(4.37\) & \((0.0095, 0.0222, 0.0583)\) & Convex; highly stable \& aids efficient training~\cite{park2019meta}. \\
        & Cross Attention & \cellcolor{red!20}\(-2.11\) & \((-0.2016, -0.0005, 0.1274)\) & Concave; fragile \& prone to optimization challenges~\cite{dauphin2014identifying}. \\
        \hline
        \multirow{3}{*}{DistilBERT} 
        & Query Projection & \cellcolor{red!20}\(-2.24\) & \((-0.1473, -0.0023, 0.0611)\) & Concave; unstable \& susceptible to faults~\cite{ghorbani2019investigation}. \\
        & Key Projection & \cellcolor{green!20}\(3.12\) & \((0.0049, 0.0150, 0.0483)\) & Convex; stable \& reliable during training~\cite{wu2018understanding}. \\
        & Output Projection & \cellcolor{green!20}\(1.75\) & \((-0.0006, 0.0150, 0.0483)\) & Convex; moderately stable with low instability~\cite{srinivas2022efficient}. \\
        \hline
    \end{tabular}
    \end{small} 
    } 
    \vspace{-1.5em}
\end{table*}

\vspace{-0.5em}
\subsection{Hessian Computation}
Although gradient-based methods show how individual parameters affect \(\mathcal{L}\), they only capture local behaviors of a model~\cite{ancona2018towards, adebayo2018sanity}. Gradients do not account for interactions among model parameters or provide a broad view of the optimization surface. As a result, they can fall short when analyzing complex curvature or unstable regions in a model's loss surface~\cite{ghorbani2019investigation}. To address these limitations, we use the Hessian matrix (\(N \times N\) matrix for \(N\) parameters in the model), the second-order derivative of \(\mathcal{L}\) with respect to \(\theta\).
\vspace{-0.5em}
\begin{equation}
\mathbf{H} = \nabla_{\theta}^2 \mathcal{L}(\theta).
\end{equation}
\vspace{-1.5em}

\(\mathbf{H}\) captures the curvature information of a loss surface and describes how the gradient changes as the parameters vary in different directions within the parameter space. The local shape of the loss function near the current parameter values can be visualized as a high-dimensional \textit{bowl}, where the \emph{eigenvalues} of \(\mathbf{H}\) quantify how steep or flat the bowl is along certain principal directions~\cite{pyhessian, ghorbani2019investigation}.

Computing \(\mathbf{H}\) directly can be expensive. We use \texttt{PyHessian}, which approximates \(\mathbf{H}\) via Hessian-vector products with \(\leq 1\%\) error on standard benchmarks~\cite{pyhessian}. We calculate two metrics~\cite{ghorbani2019investigation}: the largest eigenvalue \(\lambda_{\text{max}}\) using the Lanczos algorithm and the trace of \(\mathbf{H}\) via the Hutchinson approximation~\cite{ghorbani2019investigation}.
\vspace{-0.5em}
\begin{equation}
\lambda_{\text{max}} = \max_{i} \bigl(\lambda_i(\mathbf{H})\bigr),
\end{equation}
\vspace{-1.5em}

\noindent
\textit{Largest Eigenvalue.} \(\lambda_{\text{max}}\) represents the maximum steepness of these directions. A large positive \(\lambda_{\text{max}}\) indicates a narrow, sharply curved region on the loss surface where small parameter updates can drastically alter the gradient~\cite{battiti1992first}. Such regions tend to be more sensitive during training and pose a higher risk of unstable updates~\cite{ghorbani2019interpretation}. Approximating \(\lambda_{\text{max}}\) provides insight into the worst-case curvature in any direction of the model’s parameter space. 

\vspace{-1.0em}
\begin{equation}
\operatorname{Tr}(\mathbf{H}) = \sum_{i} \lambda_i(\mathbf{H}),
\end{equation}
\vspace{-1.0em}

\noindent
\textit{Trace of the Hessian.} \(\operatorname{Tr}(\mathbf{H})\) is the sum of all eigenvalues and reveals that the overall steepness across many directions within the parameter space. High traces imply that the loss surface is sharply curved in multiple directions, increasing the likelihood of oscillating updates and training instability~\cite{pyhessian}. 
\vspace{-0.5em}
\subsection{Attention Parameter-Layer Isolation}
\label{subsec:Attention Derivative Module}
We developed a custom module to analyze the gradients and Hessian values of our selected attention-based models. First, we isolate attention-specific parameters in each model by identifying their unique module paths in the computational graph. Then, we compute gradients and Hessian values at the parameter level using appropriate tools, such as PyHessian~\cite{pyhessian}. 

Although the Hessian-based analysis is performed at the parameter level, we also aggregate the results to the layer level to align with the architectural components (e.g., word-level attention). This aggregation allows us to interpret the findings in terms of model design, providing actionable information on fault diagnosis. By focusing on layers, we ensure that the results are directly applicable to identifying and resolving faults within the network's structure.
\vspace{-0.5em}
\subsection{Controlled Parameter Perturbation}
To simulate training faults in our deep learning models (e.g., training instabilities), we systematically introduced parameter perturbations~\cite{madhyastha2019model, pratte2013attention} into key attention components during model training. We injected noise directly into specific parameters within the training loop, leveraging PyTorch's flexibility for real-time modifications~\cite{li2024deep}.
\vspace{-1.0em}
\[\theta_{\text{perturbed}} = \theta + \alpha \delta
\] 
Here, \(\delta\) follows a Gaussian distribution (\(\delta \sim \mathcal{N}(0, I)\)), and \(\alpha\) controls the perturbation magnitude. These perturbations serve as proxies for real-world faults that disrupt the model’s learning~\cite{garcia2024effects, hua2021noise}. We focused on key attention components (e.g., word-level attention, cross-attention layers, and query/key/value projections) because of their critical role in information flow and their potential sensitivity to noise. By targeting these components, we could analyze how local faults propagate across the network, revealing fragile regions and interdependencies between layers. To evaluate model stability, we measured changes in loss, gradient norms, and Hessian properties relative to an unperturbed baseline across varying \(\alpha\) values. We conducted two experimental runs for each setup: (1) a baseline with unperturbed parameters and (2) a version with controlled perturbations. Comparing these runs allowed us to quantify the impact of perturbations on loss behavior and interdependencies. We ran our experiments in PyTorch, using standard settings recommended by the original authors~\cite{pytorch_text_classification, hmdb51_recognition, huggingface_transformers}. This setup enabled us to analyze potential instabilities without requiring separate faulty models.

\section{Results}
\textbf{Answering RQ$_1$:} We analyze the curvature of the loss surface using the Hessian method to identify \textit{fragile regions} in three attention-based neural networks: HAN, 3D-CNN, and DistilBERT. Fragile regions are areas with sharp curvatures~\cite{alain2019negative, ghorbani2019investigation}, where optimization becomes unstable. Identifying these regions is critical for pinpointing the root causes of faults (e.g., instabilities or misaligned attention), allowing us to trace fault symptoms (e.g., oscillations or divergence during training) back to specific layers or parameters.

In the HAN model, word-level attention exhibited a negative Hessian trace of \(-5.72\) (Table~\ref{tab:curvature_analysis}), indicating a concave loss surface with sharp curvatures. These concave regions make training highly sensitive to parameter changes, as shown by the steep loss curve in Fig.~\ref{fig:loss_pertubation}. In contrast, sentence-level attention showed a positive Hessian trace of \(2.86\), corresponding to a convex and stable loss surface. The flat and consistent loss curve in Fig.~\ref{fig:loss_pertubation} confirms that sentence-level attention parameters are less 
prone to training failures, making word-level attention the primary source of instability.

The 3D-CNN model showed similar findings. Spatial and temporal attention layers exhibited positive Hessian traces (\(1.92\) and \(4.37\), respectively), suggesting stable convex regions. However, the cross-attention layer demonstrated a negative Hessian trace of \(-2.11\), highlighting its fragility. This is clearly shown in Fig.~\ref{fig:loss_pertubation}, where the cross-attention loss curve is far more sensitive to perturbations compared to spatial and temporal components. These results suggest that cross-attention serves as a critical failure point during the integration of spatial and temporal streams.

For DistilBERT model, the query projection layer displayed a negative Hessian trace of \(-2.24\), placing it in a fragile concave region. In contrast, the key and output projection layers exhibited positive Hessian traces (\(3.12\) and \(1.75\)), indicating more stability (Table~\ref{tab:curvature_analysis}). Fig.~\ref{fig:loss_pertubation} supports these observations, as the query projection loss curve fluctuates significantly under perturbations, while the key and output projections remain stable.

Hessian-based curvature analysis offers deeper insights into the loss surface compared to gradients, which only measure the immediate slope. For example, gradients can highlight large updates for word-level attention parameters in the HAN model, but cannot determine whether these updates occur in stable (convex) or unstable (concave) regions. Hessian analysis goes beyond by identifying the curvature direction and quantifying its magnitude, allowing us to locate fragile regions precisely and understand their intensity. This additional information is invaluable for diagnosing the source of faults in attention-based models, as it helps trace instabilities to specific layers or parameters, allowing targeted fixes to improve model stability and performance.

\begin{table}[ht]
\centering
\vspace{-1.0em} 
\caption{Hessian Off-Diagonal Interactions in HAN Model}
\vspace{-1.0em} 
\small
\label{tab:han_hessian_excerpt}
\renewcommand{\arraystretch}{0.85} 
\setlength{\tabcolsep}{3pt} 
\begin{tabular}{c|cc|cc}
\toprule
 & \multicolumn{2}{c|}{\textbf{Word-Level}} & \multicolumn{2}{c}{\textbf{Sentence-Level}} \\
 & W$_1$ & W$_2$ & S$_1$ & S$_2$ \\
\midrule
W$_1$ & \cellcolor{gray!20}-- & 0.14 & -0.51 & -0.09 \\
W$_2$ & 0.14 & \cellcolor{gray!20}-- & \cellcolor{red!20}\textbf{-0.68} & 0.04 \\
\midrule
S$_1$ & -0.51 & \cellcolor{red!20}\textbf{-0.68} & \cellcolor{gray!20}-- & 0.08 \\
S$_2$ & -0.09 & 0.04 & 0.08 & \cellcolor{gray!20}-- \\
\bottomrule
\end{tabular}
\vspace{-1.0em} 
\end{table}

\textbf{Answering \textbf{RQ$_2$}:} We analyzed parameter interactions using the off-diagonal elements of the Hessian matrix, which quantify how changes in one parameter influence others. Here, diagonal values show self-parameter interactions (can be measured by gradients alone), while off-diagonal values capture parameter interdependencies and their mutual influences. These interactions revealed critical interdependencies within attention-based models, highlighting how faults propagate across layers. This global view of parameter relationships is invaluable for diagnosing faults and identifying their root causes.

In the HAN model, we observed significant interdependencies between word-level and sentence-level attention parameters. Table~\ref{tab:han_hessian_excerpt} shows a notable negative correlation (\(-0.68\)) between word-level parameter \(W_2\) and sentence-level parameter \(S_1\). This indicates that small perturbations in \(W_2\) can cascade to \(S_1\), destabilizing sentence-level representations. For example, when we introduced controlled perturbations to \(W_1\) and \(W_2\), the percentage of inconsistent predictions (referred to as prediction variability) increased from 5\% to 30\% on a fixed test set. 

To validate these insights for the HAN model (Table~\ref{tab:han_hessian_excerpt}), we implemented targeted interventions to stabilize the identified interdependencies. Specifically, we reduced the learning rate for \(W_1\) and \(W_2\) because these word-level attention parameters exhibited strong coupling with sentence-level parameters, as reflected by the off-diagonal Hessian magnitude of \(-0.68\). This strong coupling caused instability, as small changes in \(W_1\) and \(W_2\) propagated to \(S_1\), leading to increased prediction variability. By lowering the learning rate for word attention, we reduced the off-diagonal Hessian magnitude to \(-0.41\), which weakened the correlation and decreased prediction variability to 10\%. This intervention demonstrated improved model reliability and stability.

The 3D-CNN and DistilBERT models exhibited similar patterns of interdependence, where interactions between layers introduced instability. In 3D-CNN, although the spatial and temporal attention layers were individually stable, their interaction through the cross-attention layer amplified perturbations, making it a critical failure point during the integration of spatial and temporal streams. Similarly, in DistilBERT, off-diagonal Hessian elements revealed moderate coupling between the query projection and output projection layers. Faults in the query projection (e.g., misaligned attention focus) could propagate downstream and disrupt final predictions.

Compared to gradient-based methods, Hessian analysis provides a more complete view of parameter relationships and fault propagation. Gradients capture parameters with immediate impact on loss but miss cascading effects across interdependent components. For instance, in HAN, gradients showed large updates in \(W_2\) (norm \(\approx1.25\)) but could not explain how this instability spread to \(S_1\), which the Hessian captured through a strong off-diagonal correlation of \(-0.68\). By uncovering such hidden interdependencies, Hessian-based analysis proves valuable for diagnosing faults in attention-based models. While Hessian approximation (with \(\leq 1\%\) error~\cite{pyhessian}) is a scalable approach for moderately large models, applying it to LLMs remains challenging due to overhead, potentially requiring higher error tolerances~\cite{grosse2023studying}. 

\vspace{-0.5em}
\section{Implications}
Our findings reveal innovative ways to diagnose and address faults in attention-based models, focusing on two key aspects.

\textbf{1. Curvature Analysis:}  
Hessian-based curvature analysis offers a novel approach to identifying fragile regions in attention-based models, where sharp curvatures in the loss surface correlate with instability. Fragile regions, such as the word attention layer in HAN (\(-5.72\)) and the cross-attention layer in 3D-CNN (\(-2.11\)), highlight components prone to training failures, including oscillations and divergence. Unlike gradient analysis, which fails to capture these high-risk regions, curvature analysis pinpoints specific layers or parameters where fault symptoms originate, enabling precise interventions. Monitoring Hessian traces could act as an early warning system for instability-prone layers during training, seamlessly integrating into dynamic fault diagnosis tools like DeepFD~\cite{deepfd} and DeepLocalize~\cite{wardat2021deeplocalize} to automate the detection of fragile regions and prevent failures.

\textbf{2. Parameter Interaction Analysis:}  
Parameter interaction analysis leverages off-diagonal Hessian elements to uncover hidden interdependencies that propagate faults across layers, a critical aspect often overlooked in current literature~\cite{deepfd, wardat2021deeplocalize, zhang2021autotrainer, umlaut}. These interdependencies explain how faults in one parameter cascade through the model to destabilize other layers. In HAN (as detailed in RQ-2), a strong negative correlation (\(-0.68\)) between word attention (\(W_2\)) and sentence attention (\(S_1\)) revealed cascading instabilities that gradients could not detect. By quantifying these interactions, parameter interaction analysis traces fault symptoms (e.g., inconsistent predictions) back to their root causes (e.g., co-dependent parameters). Tools like AutoTrainer~\cite{zhang2021autotrainer} could integrate these insights to extend beyond gradient norms, enabling more robust and comprehensive fault diagnosis for complex attention-based models.

These initial results suggest that Hessian-based analysis can reveal fault sources often overlooked by traditional methods, especially in attention-based models. It complements current SE practices by facilitating more precise, model-specific fault diagnosis.

\section{Conclusion \& Future Work}
In this study, we investigate whether Hessian-based analysis can support fault diagnosis in attention-based models by addressing two research questions on identifying fragile regions (RQ\(_1\)) and revealing subtle parameter interactions (RQ\(_2\)). Our empirical results across HAN, 3D-CNN, and DistilBERT highlight the unique ability of Hessian metrics to detect sharp curvatures and expose interdependencies that gradients overlook. These findings suggest a promising direction for diagnosing and mitigating training instabilities in large-scale attention-based models. In the future, we plan to extend our work by integrating Hessian-based analysis into dynamic fault monitoring and mitigation frameworks into more diverse architectures. 

\noindent
\textbf{Acknowledgement:} This work was supported in part by the Natural Sciences and Engineering Research Council of Canada.

\balance
\clearpage
\bibliographystyle{ACM-Reference-Format}
\bibliography{references}

\end{document}